\documentclass[conference,a4paper]{IEEEtran}
\IEEEoverridecommandlockouts
\usepackage{cite}
\usepackage{amsmath,amssymb,amsfonts}
\usepackage{algorithmic}
\usepackage{graphicx}
\usepackage{textcomp}
\usepackage{xcolor}

\usepackage{algorithm}
\usepackage{multirow}
\usepackage[caption=false]{subfig}
\usepackage{mathtools}
\def\BibTeX{{\rm B\kern-.05em{\sc i\kern-.025em b}\kern-.08em
    T\kern-.1667em\lower.7ex\hbox{E}\kern-.125emX}}
\begin{document}

\title{A cGAN Ensemble-based Uncertainty-aware Surrogate Model for Offline Model-based Optimization in Industrial Control Problems}

\author{\IEEEauthorblockN{Cheng Feng}
\IEEEauthorblockA{\textit{Siemens Technology, Beijing, China} \\
cheng.feng@siemens.com}}

\maketitle

\begin{abstract}
This study focuses on two important problems related to applying offline model-based optimization to real-world industrial control problems. The first problem is how to create a reliable probabilistic model that accurately captures the dynamics present in noisy industrial data. The second problem is how to reliably optimize control parameters without actively collecting feedback from industrial systems. Specifically, we introduce a novel cGAN ensemble-based uncertainty-aware surrogate model for reliable offline model-based optimization in industrial control problems. The effectiveness of the proposed method is demonstrated through extensive experiments conducted on two representative cases, namely a discrete control case and a continuous control case. The results of these experiments show that our method outperforms several competitive baselines in the field of offline model-based optimization for industrial control.
\end{abstract}


\section{Introduction}
\label{sec:introduction}
\IEEEPARstart{T}{o} maintain the stability and efficiency of production process, it is common to use controllers ranging from proportional controllers, state estimators to advanced predictive controllers such as model predictive control (MPC) in both discrete and process manufacturing industries. However, the design procedure of these classic controllers often involves careful analysis of the process dynamics, development of abstract mathematical models, and finally derivation of control policies that meet certain design criteria. To this end, classic industrial control algorithms oftentimes suffer from sub-optimal performance, heavy requirement on domain knowledge and low scalability. With recent advances of AIoT (AI + IoT) techniques, applying data-driven methods in industrial control problems has become an exciting area of research \cite{hein2017benchmark,nian2020review}. With given conditional parameters, the hope is that a data-driven control policy can output optimized control parameters which will lead to desired control results with better production quality, higher energy efficiency, and lower air pollutant emission, etc. Nevertheless, as experimental cost is often prohibitively expensive in industrial systems, policies have to be learned from offline data without further active data collection. As a result, the reliability of learned control policies becomes a major concern which hinders their application in practice.  

Recently, many offline model-based optimization algorithms in the context of contextual bandits \cite{trabucco2021conservative} and Markov decision process \cite{yu2020mopo,kidambi2020morel} have been proposed to improve the reliability of data-driven control by constraining policies to be more conservative when learned from offline data. Implementation of offline model-based optimization typically consists of two steps: 1) Learn a surrogate model from offline data which maps from pairs of condition and control inputs into their resulting states/rewards; 2) Optimize a policy on top of the learned surrogate model via optimization techniques, for instance, genetic algorithms \cite{sivanandam2008genetic}, Bayesian optimization \cite{shahriari2015taking}, reinforcement learning (RL) \cite{sutton2018reinforcement}, with specific mechanisms to mitigate the distribution shift problem \cite{levine2020offline}. While showing promising results on games and physical simulation tasks \cite{fu2020d4rl}, their application on real-world industrial control problems is not well studied to our knowledge. Apart from the distribution shift problem, there are also other challenges for learning industrial control policies from offline data: 1) Industrial processes are often highly noisy due to reasons such as only partially observable states, imperfect sensor measurements and actuator controls. As a result, it is important to accurately capture both the aleatoric and epistemic uncertainties in the system, and ensure the learned policies only be conservative with respect to the epistemic uncertainty, but not to the aleatoric uncertainty which is inherently high in industrial systems. 2) Control results are often multi-dimensional and they are highly correlated with each other. 3) Conditional distribution of control results are often non-Gaussian. Consequently, frequently used probabilistic models such as Gaussian processes \cite{rasmussen2003gaussian} and neural networks which assume that different dimensions of the outputs are conditionally independent and follow Gaussian distributions \cite{yu2020mopo,kidambi2020morel} can cause significant errors on reward calculation. We argue that failure to tackle any one of the above challenges can bring significant risks for applying the learned control policy in practice. 

In this work, we propose a novel surrogate model which consists of two main components to achieve reliable offline model-based optimization in industrial control problems. The first component is an ensemble of conditional generative adversarial networks \cite{mirza2014conditional} (cGAN) which models the aleatoric uncertainty in the industrial system with minimum assumption on the hidden dynamics. The second is an uncertainty-aware reward penalization component to accurately avoid giving over-estimated rewards to Out-of-Distribution (OoD) inputs during the optimization process. Extensive experiments on two representative cases, a discrete control case and a continuous control case, show that our proposed method can effectively improve the reliability and performance of learned control policy compared with several competitive baselines.

\section{Background}

\subsection{Offline model-based industrial control}
We consider typical industrial control problems where there are a set of conditional variables $\mathcal{X}$ (conditional variables can only be measured, but uncontrollable), a set of control variables $\mathcal{U}$ and a set of control result variables $\mathcal{Y}$. Without loss of generality, we assume a reward function $r(\mathbf{y})$ is always known and it outputs a scalar reward value given the observed control result $\mathbf{y}$. Furthermore, there is a hidden dynamics model $p(\mathbf{y}|\mathbf{x},\mathbf{u})$ which governs the distribution of the control results given specific conditional parameters $\mathbf{x}$ and control parameters $\mathbf{u}$.

Given a historical dataset $\mathcal{D}=\{ (\mathbf{x}_t,\mathbf{u}_t,\mathbf{y}_t) \}_{t=1}^N$ in which each tuple consists of the logged conditional parameters, control parameters and control results at a discrete time step $t$ (control results can be observed at a future time $t+\delta$, but logged with time $t$ in order to align with the corresponding conditional and control parameters), the target is to learn a control policy $\pi (\mathbf{x}_t)$ which outputs the (near-) optimal control parameters at an arbitrary time $t$ that maximize the expected reward for the industrial process given the observed conditional parameters.

Furthermore, we consider two typical classes of industrial control, namely discrete control and continuous control. Specifically, in discrete control cases where control events are independent with each other, learning the optimal control policy becomes a contextual bandits or one-step optimization problem, thus we can describe the optimization problem as follows:
\begin{equation}
\label{eq:obx}
     \max_{\pi} \mathbb{E}_{\mathbf{x},\mathbf{u} \sim \pi} \Big[ \int p(\mathbf{y}|\mathbf{x},\mathbf{u}) r(\mathbf{y}) \mathrm{d}\mathbf{y} \Big]
\end{equation}where the subscript $t$ is ignored whenever it is clear that $\mathbf{x},\mathbf{u},\mathbf{y}$ share the same $t$ hereafter, and the expectation symbol is used to allow the learned control policy to be stochastic. In continuous control cases, control events are correlated. That is to say the set of conditional variables and control result variables overlap with each other, i.e., $\mathcal{X} \cap \mathcal{Y} \neq \emptyset$. For example, in typical process manufacturing scenarios the concentration of a chemical component at time step $t$ is both a control result of time step $t-1$ and a conditional parameter of time step $t$. In these cases, we can describe the optimization problem as follows: 
\begin{equation}
\label{eq:orl}
     \max_{\pi} \mathbb{E}_{\mathbf{x}_{0:T},\mathbf{u}_{0:T} \sim \pi} \Big[\sum_{t=0}^T \gamma^t \int p(\mathbf{y}_{t}|\mathbf{x}_{t},\mathbf{u}_{t}) r(\mathbf{y}_{t}) \mathrm{d}\mathbf{y}_{t} \Big]
\end{equation}where $\gamma \in (0,1)$ is a discount factor for future rewards, $T$ is a finite or infinite time horizon. 

To solve the above optimization problems, a common approach is to fit a probabilistic surrogate model via the historical dataset to approximate the dynamics model, examples are Gaussian processes \cite{deisenroth2013gaussian}, Bayesian neural networks \cite{depeweg2018decomposition} and the most commonly used ensemble of Gaussian probabilistic neural networks (GPNs) which predict the conditional distribution of outputs using a multivariate Gaussian with a diagonal covariance structure \cite{yu2020mopo,kidambi2020morel}; then the optimization problem for discrete control in Equation~\ref{eq:obx} can be solved by black-box optimization algorithms such as Bayesian optimization and genetic algorithms, and the optimization problem for continuous control in Equation~\ref{eq:orl} can be solved by model-based offline RL. However, a well-known challenge for such offline model-based optimization problems is the distribution shift problem \cite{levine2020offline} which is mainly caused by limited knowledge of the process dynamics that is covered by the historical dataset. Due to the distribution shift problem, the learned control policy from offline data often yields poor results online because it may output OoD control parameters with over-estimated rewards. To this end, offline model-based optimization algorithms often adopt conservative objective models that penalize OoD control parameters which exhibit high predictive (epistemic) uncertainty \cite{yu2020mopo,kidambi2020morel}. Apart from the distribution shift problem, the noisy, inter-dependency and non-Gaussian dynamics of control results also pose challenges for learning reliable control policies for industrial processes from offline data.

\subsection{Generative adversarial networks}
GANs \cite{goodfellow2014generative} are an example of generative models that aim to learn an estimate (i.e. $P(\mathbf{x}|\boldsymbol{\theta})$) of the data distribution $P_{\text{data}}(\mathbf{x})$, given the training samples drawn from $P_{\text{data}}(\mathbf{x})$, such that we can generate new samples from $P(\mathbf{x}|\boldsymbol{\theta})$. The general idea of GANs is to construct a game between two players. One of them is called the generator, which intends to create samples following the same distribution as the training data. The other player is the discriminator which tries to distinguish whether the samples (obtained from the generator) are real or fake. When the discriminator cannot tell apart generated and training samples, it implies that we have learned the data distribution, i.e. $P(\mathbf{x}|\boldsymbol{\theta}) \approx P_{\text{data}}(\mathbf{x})$.

The generator is simply a differentiable function $\mathbf{x} = G(\mathbf{z};\boldsymbol{\theta}_G)$ with model parameters $\boldsymbol{\theta}_G$, where the input $\mathbf{z}$ follows a simple prior distribution, such as uniform and Gaussian distribution. The discriminator is a two-class
classifier, $D(\mathbf{x}; \boldsymbol{\theta}_D)$ often designed as a neural network to
output class probability between 0 and 1. GAN aims to solve the following min-max optimization problem:

\begin{footnotesize}
\begin{eqnarray*}
\min_{\boldsymbol{\theta}_G} \max_{\boldsymbol{\theta}_D}  \mathbb{E}_{\mathbf{x} \sim P_{\text{data}}(\mathbf{x})}[\log D(\mathbf{x}; \boldsymbol{\theta}_D)] +\mathbb{E}_{\mathbf{z}\sim P(\mathbf{z})}[\log(1-D(G(\mathbf{z}; \boldsymbol{\theta}_G)))]
\end{eqnarray*}\end{footnotesize}An alternative update of $\boldsymbol{\theta}_G$ and $\boldsymbol{\theta}_D$ by stochastic gradient descent can be adopted to solve this problem. Then we can use the optimized $\mathbf{x} = G(\mathbf{z}; \boldsymbol{\theta}_G^*)$ to generate new samples through the random input $\mathbf{z}$, which provides a Monte-Carlo approach to estimate the data distribution $P_{\text{data}}(\mathbf{x})$.

\subsection{Deep ensemble for uncertainty quantification}
From a Bayesian viewpoint, the uncertainty of a model at a specific input can be decomposed into two categories \cite{gal2016uncertainty}: 
\begin{itemize}
\item Aleatoric uncertainty: uncertainty caused by inherent variation of the system in that region -- collecting more data would not reduce this type of uncertainty.
\item Epistemic uncertainty: uncertainty due to lack of knowledge about that region -- can be reduced if collecting more data from that region of input space.
\end{itemize}
A promising approach for quantifying uncertainty is through deep ensembles~\cite{lakshminarayanan2017simple}. Concretely, deep ensembles are ensembles of deep probabilistic neural networks trained on the same dataset with different random parameter initialisations and data shuffling. The ensemble members can be interpreted as samples from different modes of the Bayesian parameter posterior $P(\boldsymbol{\theta}|\mathcal{D})$, which allows their predictions to be more diverse compared to unimodal approximate Bayesian approaches.

Following the Bayesian interpretation of Deep Ensembles \cite{xia2022usefulness}, we obtain an approximate posterior predictive distribution by averaging over the outputs of the individual ensemble members:
\begin{eqnarray*}
P(\mathbf{y}|\mathbf{x},\mathcal{D}) \approx P(\mathbf{y}|\mathbf{x},\boldsymbol{\Theta}) =\frac{1}{M} \sum_{m=1}^M P(\mathbf{y}|\mathbf{x},\boldsymbol{\theta}^{(m)})
\end{eqnarray*} where $M$ is the number of ensemble members. Furthermore, we can decompose uncertainty at input $\mathbf{x}$ measured by the predictive entropy in the following form:
\begin{eqnarray*}
\mathcal{U}(\mathbf{x}) &\approx& \underbrace{\mathcal{H}\big( P(\mathbf{y}|\mathbf{x},\boldsymbol{\Theta}) \big) - \frac{1}{M}\sum_{m=1}^M \mathcal{H}\big( P(\mathbf{y}|\mathbf{x},\boldsymbol{\theta}^{(m)}) \big)  }_{\mathclap{\text{Epistemic Uncertainty}}} \\
&& + \underbrace{\frac{1}{M}\sum_{m=1}^M \mathcal{H}\big( P(\mathbf{y}|\mathbf{x},\boldsymbol{\theta}^{(m)}) \big) }_{\mathclap{\text{Aleatoric Uncertainty}}}
\end{eqnarray*}The epistemic uncertainty, when approximated using above equation, measures the diversity of the ensemble, as it is the mean KL-divergence between the predictive distributions of the ensemble members and the ensemble overall:
\begin{eqnarray*}
\mathcal{H}\big( P(\mathbf{y}|\mathbf{x},\boldsymbol{\Theta}) \big) - \frac{1}{M}\sum_{m=1}^M \mathcal{H}\big( P(\mathbf{y}|\mathbf{x},\boldsymbol{\theta}^{(m)}) \big) \\
=  \frac{1}{M}\sum_{m=1}^M KL[P(\mathbf{y}|\mathbf{x},\boldsymbol{\theta}^{(m)}) || P(\mathbf{y}|\mathbf{x},\boldsymbol{\Theta}) ]
\end{eqnarray*} As a result, the diversity of deep ensemble members are widely utilized to quantify epistemic uncertainty, and the average uncertainty of deep ensemble members are utilized to quantify aleatoric uncertainty. However, recently there are also arguments that deep ensemble diversity does not necessarily indicates epistemic uncertainty \cite{xia2022usefulness,abe2022deep,xia2023window}.

\section{Proposed surrogate Model}
In this section we present proposed cGAN ensemble-based surrogate model which can accurately capture both the aleatoric uncertainty and the epistemic uncertainty in general industrial systems, and thus can achieve accurate and reliable reward calculation for arbitrary conditional and control inputs.

\subsection{cGAN ensemble-based probabilistic model}
\label{sec:surrogate}

Given a historical dataset $\mathcal{D}=\{ (\mathbf{x}_t,\mathbf{u}_t,\mathbf{y}_t) \}_{t=1}^N$, the generator and discriminator of our cGAN models are designed as follows:

\emph{Generator} ($G$): The input of the generator includes a noise vector ($\mathbf{z}$) consisting of randomly generated float numbers between 0 and 1, and two condition vectors which are the conditional parameters ($\mathbf{x}$) and the control parameters ($\mathbf{u}$). The output of the generator is the vector of control results $\tilde{\mathbf{y}}$. We define the generator as a function $\tilde{\mathbf{y}}=G(\mathbf{z}|\mathbf{x},\mathbf{u})$. The objective of the generator is to generate simulated control results that cannot be distinguished by the discriminator, denoted as follows:
\begin{equation}
    \min_{G }\mathbb{E}_{(\mathbf{x},\mathbf{u}) \sim p(\mathbf{x},\mathbf{u}), \mathbf{z} \sim p(\mathbf{z})} -D(G(\mathbf{z} | \mathbf{x},\mathbf{u}) | \mathbf{x},\mathbf{u}) \nonumber
\end{equation}where $p(\mathbf{x},\mathbf{u})$ is the empirical joint distribution of conditional and control parameters in the historical data $\mathcal{D}$, $p(\mathbf{z})$ is a multivariate distribution where each dimension follows an independent uniform distribution between 0 and 1.

\emph{Discriminator} ($D$): The discriminator also has two part of inputs. The first part is the two condition vectors $\mathbf{x}$ and $\mathbf{u}$ as the same as the counterpart in the generator. The second part is either a generated control result sample $\tilde{\mathbf{y}}$ or a real control result sample ($\mathbf{y}$). The output of the discriminator is a scalar value. A lower value of the output indicates the discriminator gives a higher likelihood to the control result sample as a generated one. The objective of the discriminator is denoted as follows:
\begin{eqnarray*}
    \min_{D} &&\mathbb{E}_{(\mathbf{x},\mathbf{u}) \sim p(\mathbf{x},\mathbf{u}),\mathbf{z} \sim p(\mathbf{z})} D(G( \mathbf{z} | \mathbf{x},\mathbf{u}) | \mathbf{x},\mathbf{u}) \\
    && -\mathbb{E}_{(\mathbf{x},\mathbf{u},\mathbf{y}) \sim p(\mathbf{x},\mathbf{u},\mathbf{y})} D( \mathbf{y} | \mathbf{x},\mathbf{u}) \\
    &&+ \lambda \mathbb{E}_{(\mathbf{x},\mathbf{u},\mathbf{y}') \sim p(\mathbf{x},\mathbf{u},\mathbf{y}')} \Big[(|| \frac{\Delta D(\mathbf{y}'|\mathbf{x},\mathbf{u})}{\Delta \mathbf{y}'} ||-1)^2 \Big]
\end{eqnarray*}
where the first term minimizes the outputs for simulated samples; the second term maximizes the outputs for real samples; the third term is the gradient penalty loss following Wasserstein GAN~\cite{gulrajani2017improved} in which $\lambda$ is a hyperparameter often takes the value 10, $\mathbf{y}'$ samples uniformly along straight lines between pairs of real and simulated control results with given conditional and control parameters.   

Furthermore, we train an ensemble of $M$ cGAN models on the historical dataset. Each cGAN model is trained with randomized initialization of parameters along with random shuffling of the training samples, but with the same structure and hyperparameters. After the $M$ cGAN models are trained, we can generate a large number of control result samples using the generators and then utilize a Monte-Carlo approach to calculate the expect reward given any conditional and control parameters. Concretely, our proposed probabilistic model $r(\mathbf{x},\mathbf{u})$ for expected reward calculation is given as follows
\begin{small}
\begin{eqnarray}
\label{eq:r_plain}
r(\mathbf{x},\mathbf{u}) =   \frac{1}{M\cdot N}\sum_{i=1}^{M}  \sum_{j=1}^N  r\big( \tilde{\mathbf{y}}_i^j \big)
\end{eqnarray}\end{small}where $N$ is a large integer number, $\tilde{\mathbf{y}}_i^j=G_{_i} (\mathbf{z}^j|\mathbf{x},\mathbf{u})$ in which $G_{_i}$ is a trained generator model in the ensemble and $\mathbf{z}^j\sim p(\mathbf{z})$.

The main benefit of the proposed probabilistic model is that we leverage a sampling-based approach which can effectively model any shape of conditional distribution for control results while naturally capturing the dependencies between different dimensions of control results. In this way, we introduce minimum bias while capturing the aleatoric uncertainty in the noisy industrial system, thus accurate reward calculation can be achieved.

\begin{figure}
\centering
\subfloat[]{\includegraphics[width=.24\textwidth]{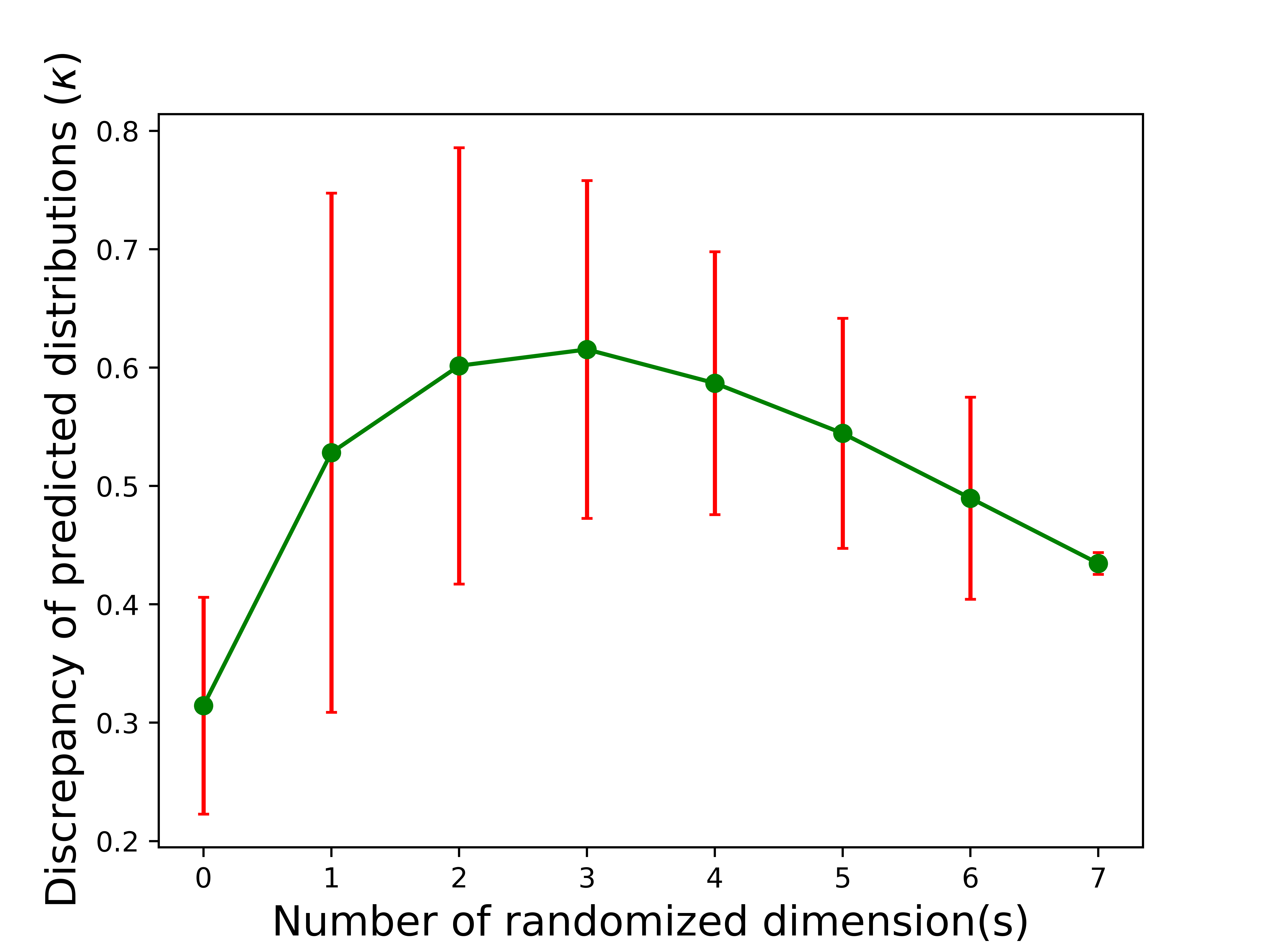}\label{fig:dim_vs_disc}}
\subfloat[]{\includegraphics[width=.24\textwidth]{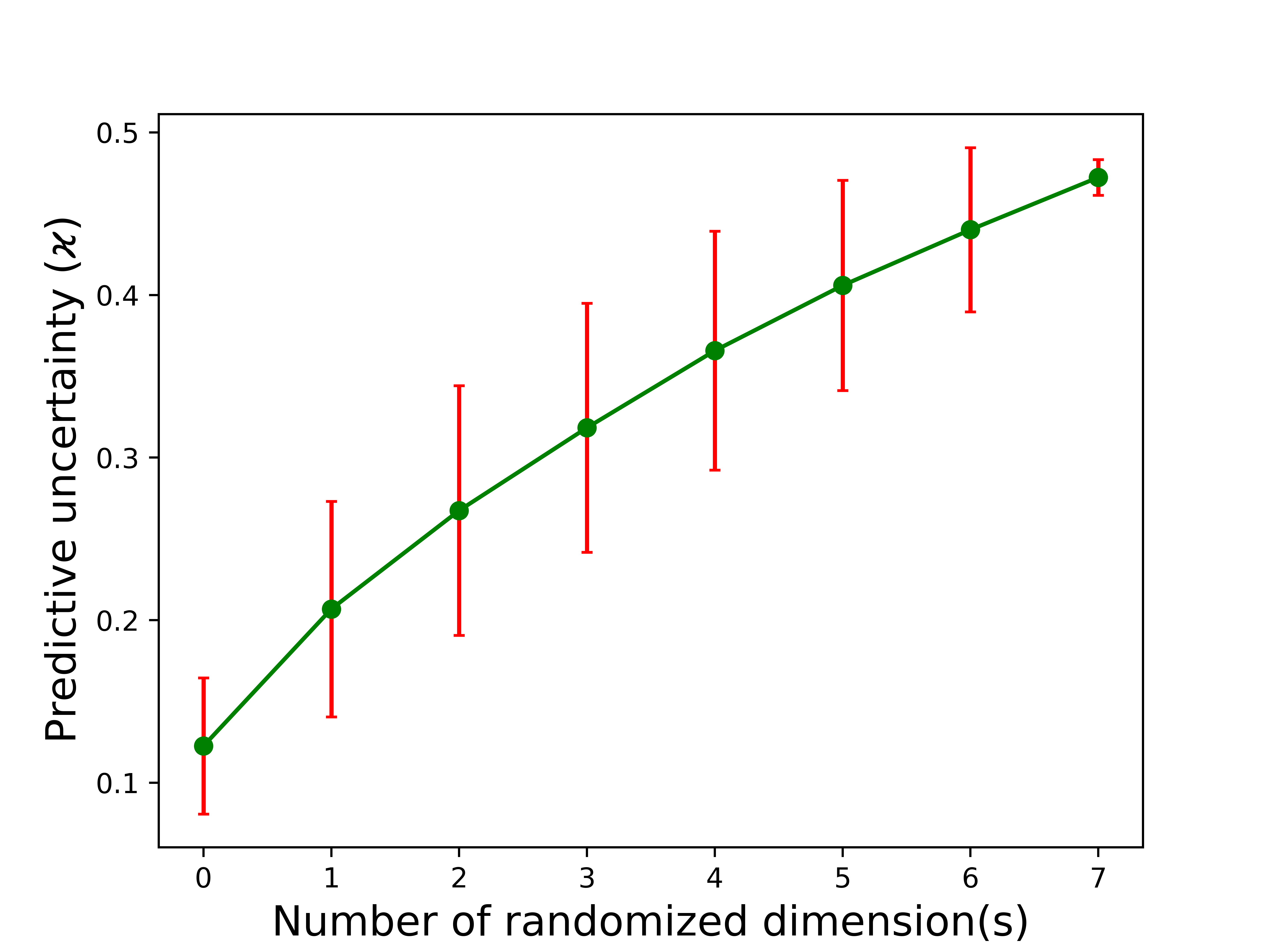}\label{fig:dim_vs_uncertainty}}
\caption{(a) the mean and standard deviation of the discrepancies between distributions of generated control results with different number of randomized dimensions for the inputs of the cGAN ensemble; (b) the mean and standard deviation of the amount of uncertainty for generated control results with different number of randomized dimensions for the inputs of the cGAN ensemble.}
\label{fig:qualitative1}
\end{figure}

\subsection{Uncertainty-aware reward penalization}
We then introduce the uncertainty-aware reward penalization component for avoiding giving over-estimated rewards to OoD inputs during the optimization process. Notably, over-estimated rewarding for OoD inputs occurs largely due to over-confident predictions are made by the surrogate model when the optimization algorithm is exploring the regions with high epistemic uncertainty. According to previous studies \cite{lakshminarayanan2017simple}, deep ensembles will generate predictions with higher uncertainty on OoD inputs because the predictions made by different models within an ensemble are likely to be diverged. Utilizing this observation, existing works propose to use either the amount of predictive uncertainty \cite{yu2020mopo} or the discrepancy between the predicted distributions \cite{kidambi2020morel} of deep ensembles to penalize rewards for avoiding assigning over-estimated rewards to OoD inputs. However, we find both methods have defects when applied in industrial control problems. On one hand, penalizing rewards with higher predictive uncertainty may lead to over-conservative policies because non-OoD inputs with inherently high aleatoric uncertainty will be incorrectly penalized. On the other hand, we observe that discrepancy between the predicted distribution of the models is not monotonically increasing when the input sample is more likely to be an outlier. In fact, we find that the discrepancy starts to decrease at some point where the input sample is sufficiently far away from the historical data distribution (see evidence from figure~\ref{fig:dim_vs_disc}). Our conjecture is that when the level of epistemic uncertainty is excessively high, it poses challenges in achieving diversity within an ensemble where each member exhibits high level of uncertainty. This finding is also consistent with the recent argument that deep ensemble diversity does not necessarily indicates epistemic uncertainty as discussed in \cite{xia2022usefulness,abe2022deep} . As a result, only penalizing rewards based on the discrepancy between predicted distributions cannot avoid giving over-estimated rewards to OoD inputs which are sufficiently far away from the historical data distribution. 

To overcome the above problems, we propose the following cGAN ensemble-based surrogate model with an uncertainty-aware reward penalization component:
\[ r_p(\mathbf{x},\mathbf{u}) =
  \begin{cases}
    (1-\kappa) \cdot r(\mathbf{x},\mathbf{u})       & \quad \text{if } \varkappa \leq \epsilon \wedge r(\mathbf{x},\mathbf{u}) > 0\\
    (1+\kappa) \cdot r(\mathbf{x},\mathbf{u})       & \quad \text{if }  \varkappa \leq \epsilon \wedge  r(\mathbf{x},\mathbf{u}) \leq 0\\
    c  & \quad \text{if } \varkappa > \epsilon
  \end{cases}
\]where $\kappa = \left(\!
    \begin{array}{c}
      M \\
      2
    \end{array}
  \!\right)^{-1} \sum_{i=1}^M \sum_{j=i+1}^M H^2_{i,j}$ is the mean squared Hellinger distance (we calculate the squared Hellinger distance via a closed form expression by assuming the generated control results follow multivariate Gaussian distributions) between the distributions of control results generated by the $M$ cGANs, and $\kappa \in (0,1)$; $\varkappa = M^{-1}\sum_{i=1}^M||\boldsymbol{\Sigma}_i||_F$ is the mean Frobenius norm of the covariance matrices of the generated control results by the $M$ cGANs; $\epsilon$ is a user-defined threshold value which can be set to, e.g., the maximum value of $\varkappa$ calculated on input samples in a validation dataset; $c$ is a constant which denotes an extremely low value to rewards. Specifically, we use $\varkappa$ to measure the uncertainty of generated control results, when it is larger than a threshold value, the epistemic uncertainty will overwhelm, thus we assign an extremely low reward to the input sample to avoid the risk of over-estimated reward estimation. When $\varkappa$ is lower than the threshold value, we penalize rewards only with epistemic uncertainty that is captured by the discrepancy between distributions of generated control results. In this way, the reward penalization component can accurately avoid assigning over-estimated rewards to OoD input samples without leading to over-conservative policies. 
  
To give an intuition about the motivation of our reward penalization component, we train an ensemble of five cGANs on the a real world industrial dataset (details to be seen in Section~\ref{sec:discrete_exp}). Moreover, we replace $n$ dimension(s) of the logged conditional and control inputs in the data with randomly generated values to mimic OoD inputs. With more randomized dimensions, the input sample is more likely to be an outlier. For each input, we generate a large number of control result samples via the five cGANs and measure the discrepancy between the distributions of generated control results of the five cGANs ($\kappa$) and the amount of predictive uncertainty ($\varkappa$). We plot the mean and standard deviation of $\kappa$ and $\varkappa$ with different value of $n$ in figure~\ref{fig:dim_vs_disc} and figure~\ref{fig:dim_vs_uncertainty} respectively. As can be seen in Figure~\ref{fig:dim_vs_disc}, the mean value of $\kappa$ increases along with $n$ when $n\leq 3$, however, it starts to decrease when $n>3$. Therefore, only penalizing rewards with the discrepancy between predicted distributions of control results can potentially assign over-estimated rewards to OoD inputs which are far away from the historical data distribution. However, since the predictive uncertainty will increase as the input is farther away from the historical data distribution as illustrated in Figure~\ref{fig:dim_vs_uncertainty}, we can effectively avoid assigning over-estimated rewards to those OoD inputs by checking whether the corresponding predictive uncertainty is larger than a threshold value beyond which the epistemic uncertainty overwhelms.

\section{Experiment on Discrete Control}
\label{sec:discrete_exp}
In this section, we present the experiment in which we applied our proposed surrogate model to optimize the control policy in a discrete control case: the production process of a worm gear production line. Specifically, worm gears, an important component in the steering system of automobiles, are produced using computer numerical control (CNC) machines. Our target is to improve the production quality of worm gears by optimizing the control of CNC machining parameters. More specifically, there are 3 critical conditional parameters and 4 critical control parameters that can significantly impact the defective rate of produced worm gears. Every produced worm gear will go through an automatic quality check, in which 7 attributes are checked to decide whether it is a qualified product. Before our learned policy is deployed, the 4 critical control parameters are manually tuned by the operators according to their personal experience.

In our experiment, we collected a historical dataset from the worm gear production line with 19760 production records. Each production record consists of the 3 conditional parameters (recorded ambient environment factors during production), 4 control parameters (CNC machining parameters) and 7 control result variables (the attributes of produced worm gears for quality check). The reward for a qualified production is set to 1, otherwise the reward is set to 0.

\subsection{Quantitative evaluation}
We utilize the optimization framework in Equation~\ref{eq:obx} to search for the optimal control parameters. For simplicity, given observed conditional parameters $\mathbf{x}$, we can convert the optimization problem as follows:
\begin{equation}
\label{eq:bayesopt}
     \mathbf{u}^* = \arg\max_{\mathbf{u} \in \mathcal{A}} r_p(\mathbf{x},\mathbf{u})
\end{equation}where $\mathcal{A}$ is the value space of control parameters. Technically, we derive the (near) optimal control parameters $\mathbf{u}^*$ via Bayesian optimization.

We consider the following four baselines to demonstrate the benefits of our proposed framework:

\textbf{Baseline1}: We ignore the epistemic-uncertainty-penalized function to evaluate rewards. This means that we replace $r_p(\mathbf{x},\mathbf{u})$ in Equation~\ref{eq:bayesopt} with $r(\mathbf{x},\mathbf{u})$ as defined in Equation~\ref{eq:r_plain}.

\textbf{Baseline2}: We replace ensemble of cGANs by ensemble of GPNs and use the same epistemic-uncertainty-penalized function to evaluate rewards. This means that we replace $\tilde{\mathbf{y}}^j_i$ in Equation~\ref{eq:r_plain} by samples from the predicted distributions of control results by GPNs. Specifically, GPNs output a Gaussian distribution over the control results with a diagonal covariance matrix: $p(\mathbf{y}|\mathbf{x},\mathbf{u}) \approx \mathcal{N}\big( {\boldsymbol{\mu}_{\theta}(\mathbf{x},\mathbf{u}),\boldsymbol{\Sigma}_{\phi}(\mathbf{x},\mathbf{u})} \big)$. We can generate control result samples by sampling from the predicted Gaussian distribution. 

\textbf{Baseline3}: We use the MOReL \cite{kidambi2020morel} style function to penalize rewards. More specifically, let disc$(\mathbf{x},\mathbf{u})=\max_{i,j} H^2_{i,j}$ be the maximum discrepancy for the predicted distributions of control results,
we replace $r_p(\mathbf{x},\mathbf{u})$ with:
\[f(\mathbf{x},\mathbf{u})
  \begin{cases}
    r(\mathbf{x},\mathbf{u})       & \quad \text{if } \text{disc}(\mathbf{x},\mathbf{u}) \leq \text{threshold}\\
    0       & \quad \text{otherwise}
  \end{cases}
\]where the threshold is a user-defined hyperparameter. As can be seen, MOReL penalizes rewards only based on discrepancy of predicted distributions of control results.

\textbf{Baseline4}: We use the MOPO \cite{yu2020mopo} style function to penalize rewards. We replace $r_p(\mathbf{x},\mathbf{u})$ with $g(\mathbf{x},\mathbf{u})$ such that:
\[ g(\mathbf{x},\mathbf{u}) = r(\mathbf{x},\mathbf{u}) - \lambda \max_{i=1,...,M} ||\boldsymbol{\Sigma}_i||_F
\]where $\lambda$ is a user-defined hyperparameter. As can be seen, MOPO penalizes rewards only based on the amount of uncertainty for predicted distributions of control results.

\subsubsection{Evaluation metrics and results}
The learned policy by our proposed method has been deployed on the production line and has increased the qualified rate of productions by $9\%$ which is rather significant. To compare the performance with the baselines, the ideal way is to run the policies learned by different methods on the production line for sufficiently long time and compare the average reward (qualified rate) of production using different policies. However, it is infeasible to do so due to technical and safety issues. As a result, we use off-policy evaluation (OPE) methods \cite{dudik2014doubly,hanna2019importance} to compare the performance of different policies. 

Specifically, we firstly split the historical data into a training set and a testing set at a 4:1 ratio. Then, we train ensemble of cGANs and GPNs on the training data and apply OPE on the testing data. Concretely, we compare the metrics calculated by the following two commonly used OPE methods: Weighted Importance Sampling (WIS) and Doubly Robust (DR) estimator. Since the WIS estimator and DR estimator are
based on two basic estimators, Direct Method (DM) and Importance Sampling (IS), we will introduce the implementation details of the four methods as follows:

\textbf{Direct Method (DM)}: The DM estimator forms an estimate $\hat{r}(\mathbf{x}_k,\mathbf{u}^*_k)$ of the expected reward given the conditional and control parameters, the expected reward of the proposed policy is then estimated by:
\begin{equation}
    \hat{V}^{\pi}_{\text{DM}} = \frac{1}{n}\sum_{k=1}^n \hat{r}(\mathbf{x}_k,\mathbf{u}^*_k) \nonumber
\end{equation}where $n$ is the number of production records in the testing data. We train a multiple layer perception (MLP) network on the training data as the predictive model to estimate $\hat{r}(\mathbf{x}_k,\mathbf{u}^*_k)$. DM estimator suffers from high bias since it highly relies on the accuracy of the predictive model.

\textbf{Importance Sampling (IS)}: The IS estimator evaluates the average reward of a policy by fixing the distribution mismatch between logging controls and policy controls:
\begin{equation}
    \hat{V}^{\pi}_{\text{IS}} = \frac{1}{n}\sum_{k=1}^n \frac{p_{\pi}(\mathbf{u}_k|\mathbf{x}_k)}{p_{\text{log}}(\mathbf{u}_k|\mathbf{x}_k)} r(\mathbf{y}_k) \nonumber
\end{equation}where $p_{\text{log}}(\mathbf{u}_k|\mathbf{x}_k)$ is the probability of taking control parameters $\mathbf{u}_k$ given conditional parameters $\mathbf{x}_k$ for the logging policy, $p_{\pi}(\mathbf{u}_k|\mathbf{x}_k)$ is the corresponding probability for the learned policy. Similar to \cite{hanna2019importance}, we estimate $p_{\text{log}}(\mathbf{u}_k|\mathbf{x}_k)$ by fitting a GPN on the historical dataset. $p_{\pi}(\mathbf{u}_k|\mathbf{x}_k)$ is estimated by fitting a Gaussian distribution for 10 output control parameters generated by Bayesian optimization each with different random seeds. The IS estimator suffers from high variance especially when the log policy differs significantly from the policy to be evaluated \cite{dudik2011doubly}.

\textbf{Weighted Importance Sampling (WIS)}: Let $w_k=\frac{p_{\pi}(\mathbf{u}_k|\mathbf{x}_k)}{p_{\text{log}}(\mathbf{u}_k|\mathbf{x}_k)}$ be the importance weight, the WIS estimator evaluates the average reward of a policy as follows:
\begin{equation}
    \hat{V}^{\pi}_{\text{WIS}} =  \frac{\sum_{k=1}^n w_k r(\mathbf{y}_k)}{\sum_{k=1}^n w_k} \nonumber
\end{equation}The WIS estimator is often more stable in practice than the IS estimator.

\textbf{Doubly Robust (DR) estimator}: The DR estimator takes the advantage of both the DM estimates and the importance weights to provide a more robust evaluation of the expected reward for a proposed policy. Specifically, we adopt the DR estimator in \cite{dudik2014doubly} as shown below: 
\begin{equation}
    \hat{V}^{\pi}_{\text{DR}} =  \frac{1}{n}\sum_{k=1}^n \big[ \hat{r}(\mathbf{x}_k,\mathbf{u}^*_k) + w_k \big( r(\mathbf{y}_k)-\hat{r}(\mathbf{x}_k,\mathbf{u}_k) \big) \big] \nonumber
\end{equation}where $\hat{r}(\mathbf{x}_k,\mathbf{u}_k)$ is the expected reward for a log control using the predictive model in the DM estimator.

In Table~\ref{tab:ope}, we present the calculated metrics for different policies using the WIS and DR methods. As can be seen, the control policy learned by our proposed method achieves the best metric using both methods. 

\begin{table}[!t]
	\renewcommand{\arraystretch}{1.3}
	\caption{The calculated OPE metrics for different policies in the discrete control case}
	\centering
	\label{tab:ope}
	\resizebox{\columnwidth}{!}{
		\begin{tabular}{lccccc}
\hline
 &   Baseline1 & Baseline2 &Baseline3 & Baseline4&  Ours\\
\hline
WIS & 0.85  & \textbf{0.86} & 0.80  & 0.75 & \textbf{0.86}  \\
DR &  0.80 &  0.61  & 0.62 & 0.84 & \textbf{0.94} \\
\hline
\end{tabular}
	}
\end{table}

\subsection{Qualitative evaluation}
In the qualitative experiments, we aim to explain why our proposed method outperforms the baseline methods by answering two questions: 1) Is it really beneficial to use cGANs instead of the commonly used GPNs in the surrogate model? 2) Is it really necessary to use both the discrepancy between predicted distributions and the amount of predictive uncertainty to penalize rewards?

\begin{figure}
\begin{center}
\centerline{\includegraphics[width=.48\textwidth]{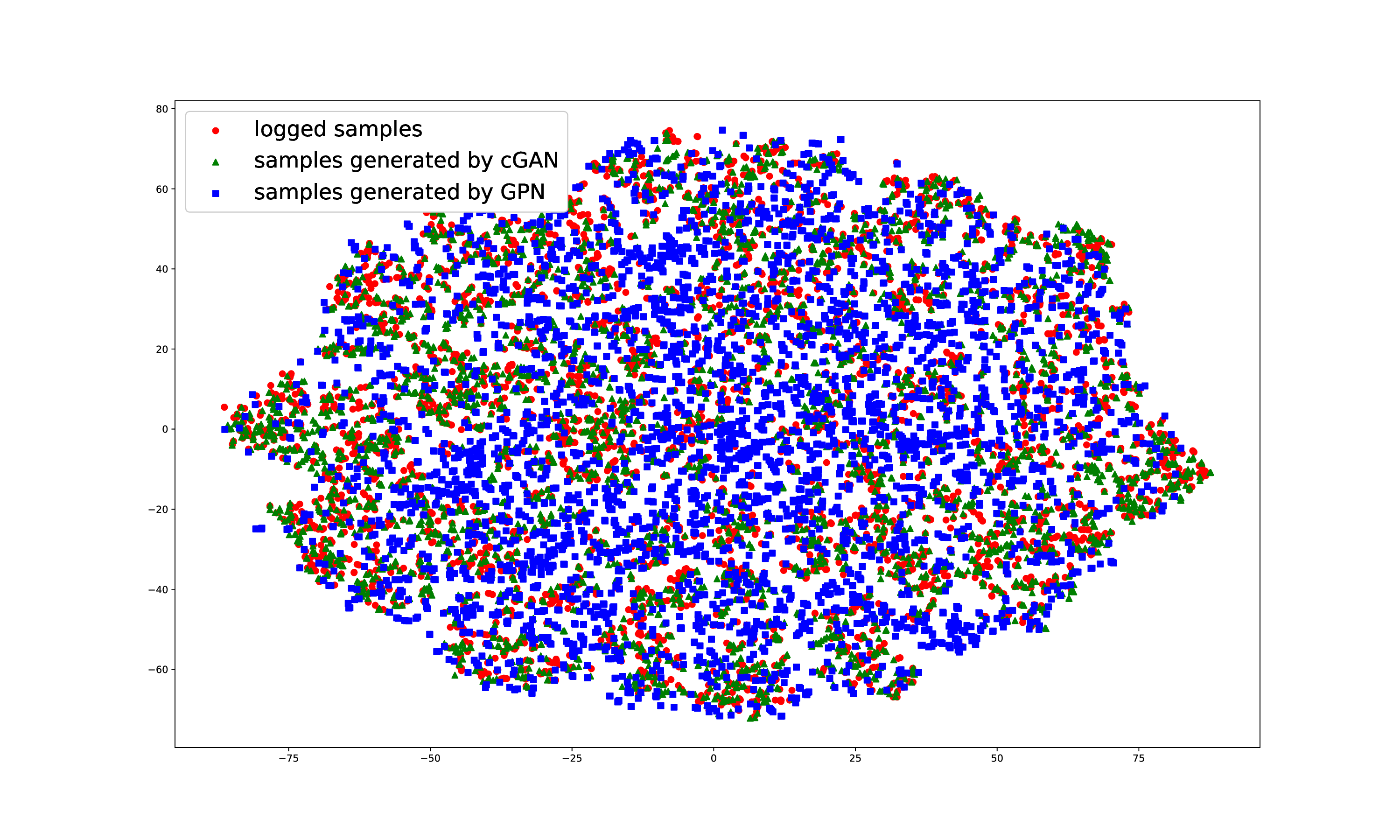}}
\caption{T-SNE 2D visualization for real and generated control result samples}
\label{fig:2d}
\end{center}
\end{figure}

To answer the first question, we use the trained cGANs and GPNs to generate equivalent number of control result samples using the logged conditional and control parameters in the testing set. We use t-SNE \cite{van2008visualizing} to visualize the logged and generated control result samples on a 2D map in Figure~\ref{fig:2d}. As can be seen, the logged control result samples have a much closer pattern with the samples generated by the cGAN models than the ones generated by the GPN models. This is because the cGAN models can reflect the true dynamics of control results by capturing the dependencies between different dimensions of the control results without the assumption for the shape of underlying distribution. Therefore, using cGANs as the probabilistic model is much more reliable than GPNs.

\begin{figure}
\centering
\subfloat[]{\includegraphics[width=.15\textwidth]{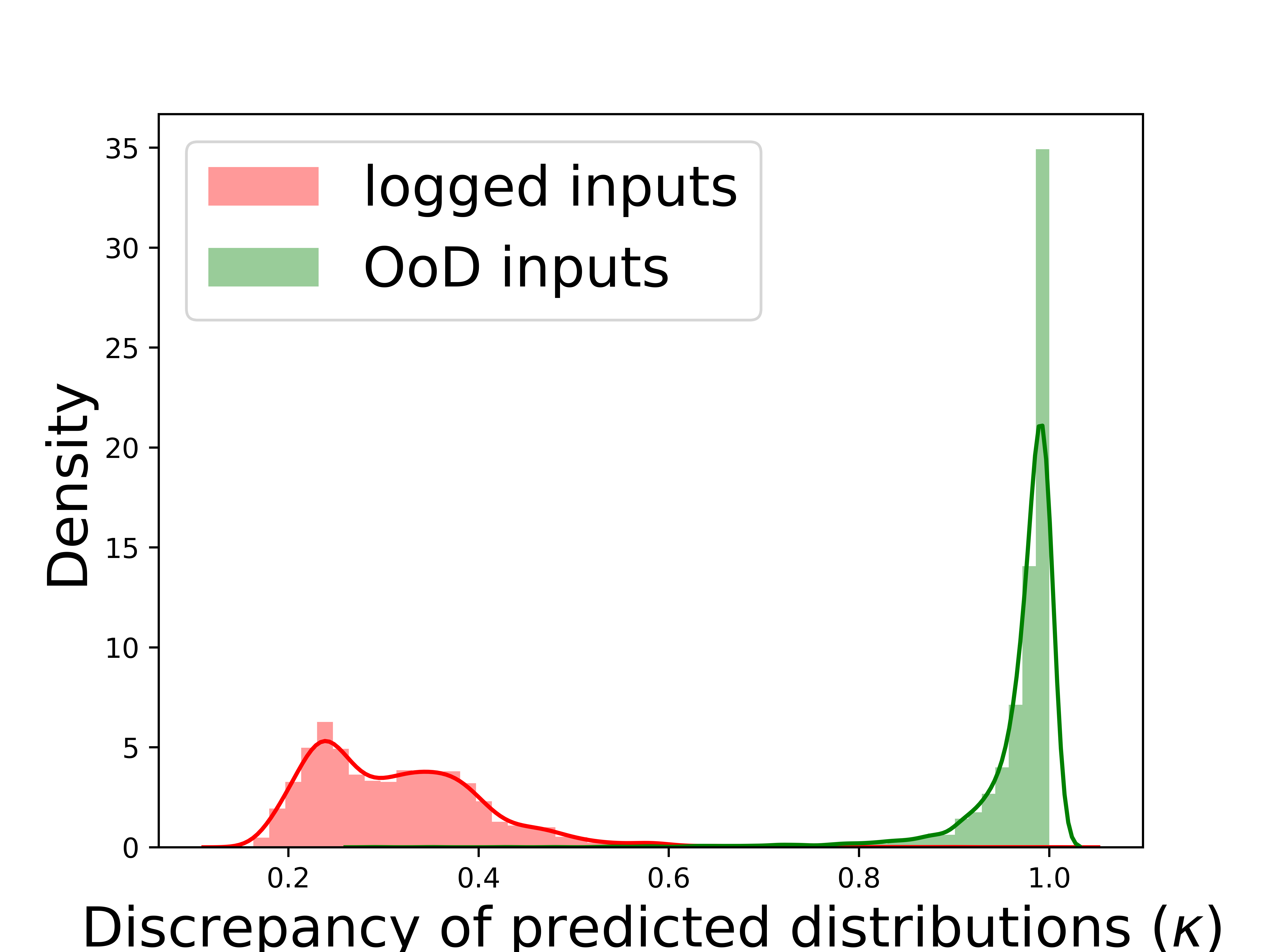}\label{fig:hist_divs2}}
\subfloat[]{\includegraphics[width=.15\textwidth]{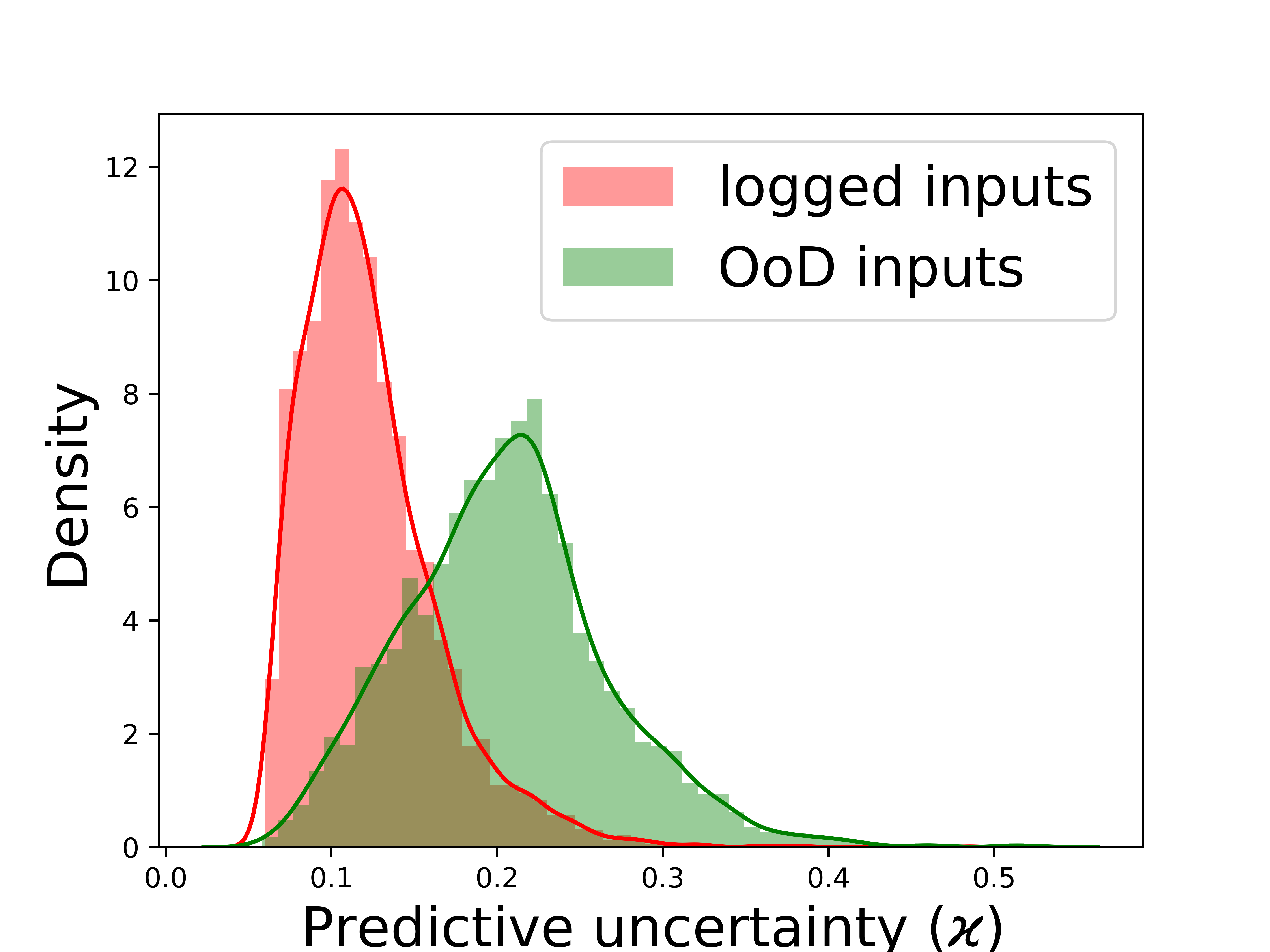}\label{fig:hist_norms2}}
\subfloat[]{\includegraphics[width=.15\textwidth]{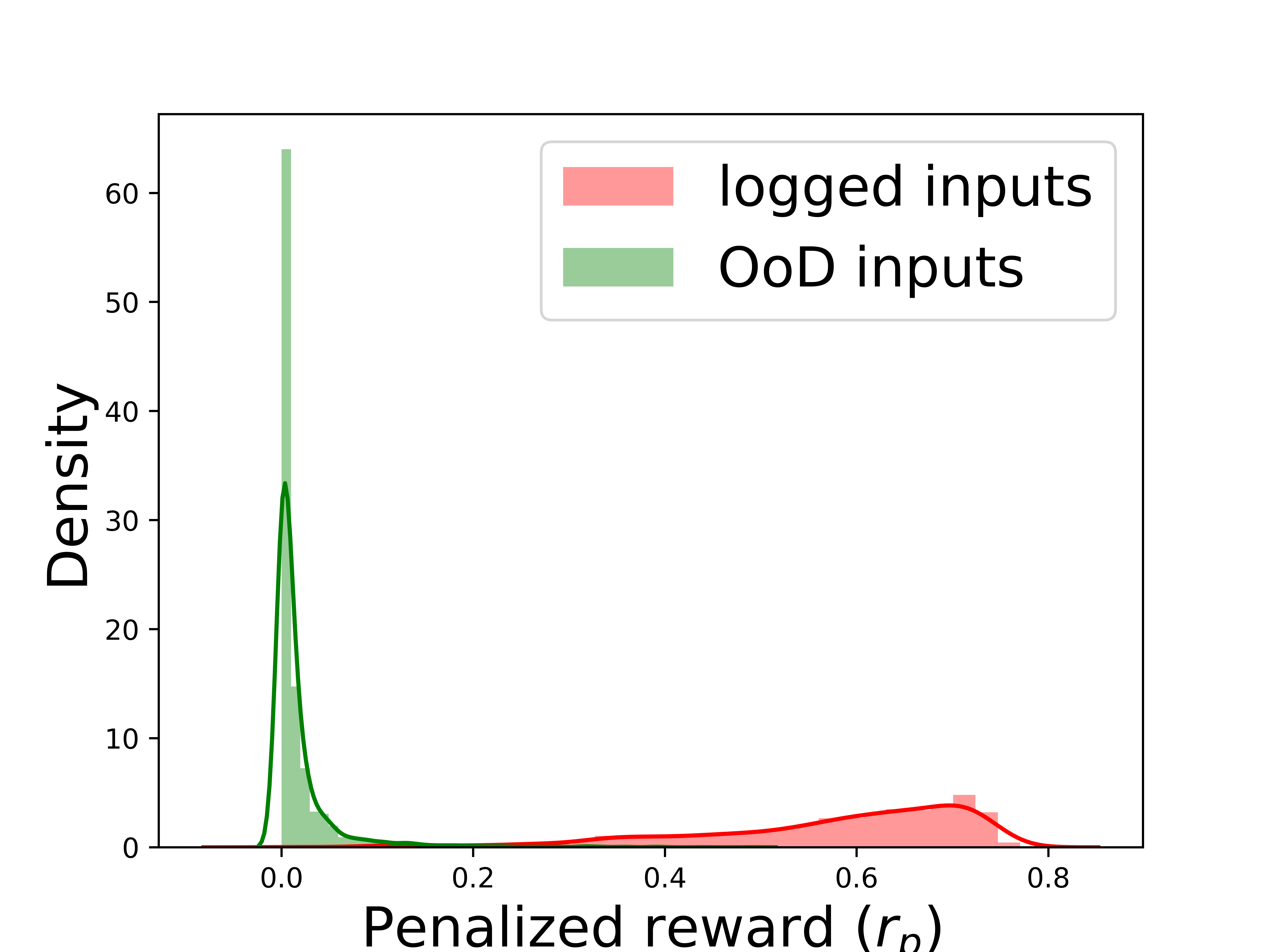}\label{fig:hist_rewards2}} \hfill
\subfloat[]{\includegraphics[width=.15\textwidth]{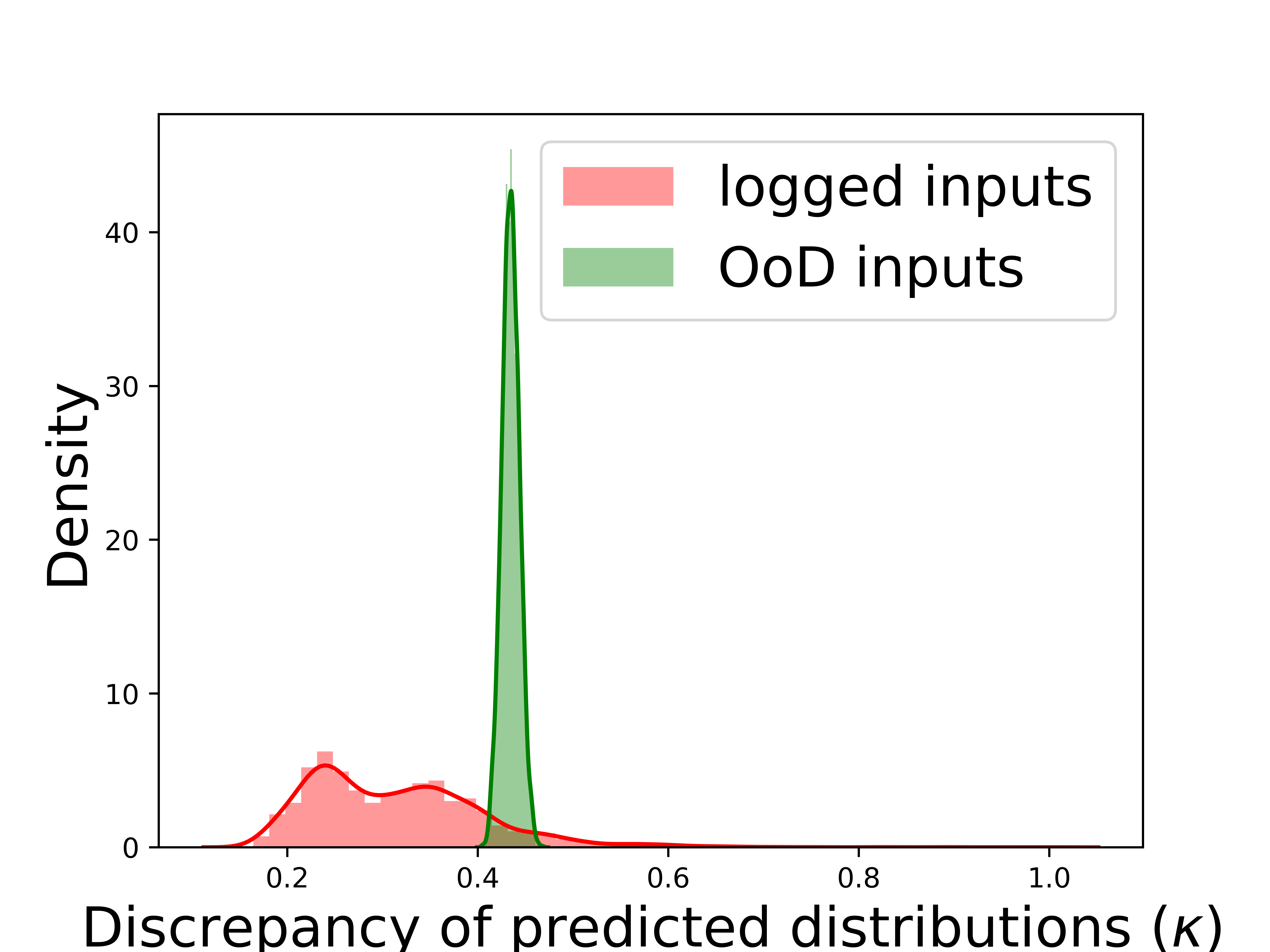}\label{fig:hist_kappa}}
\subfloat[]{\includegraphics[width=.15\textwidth]{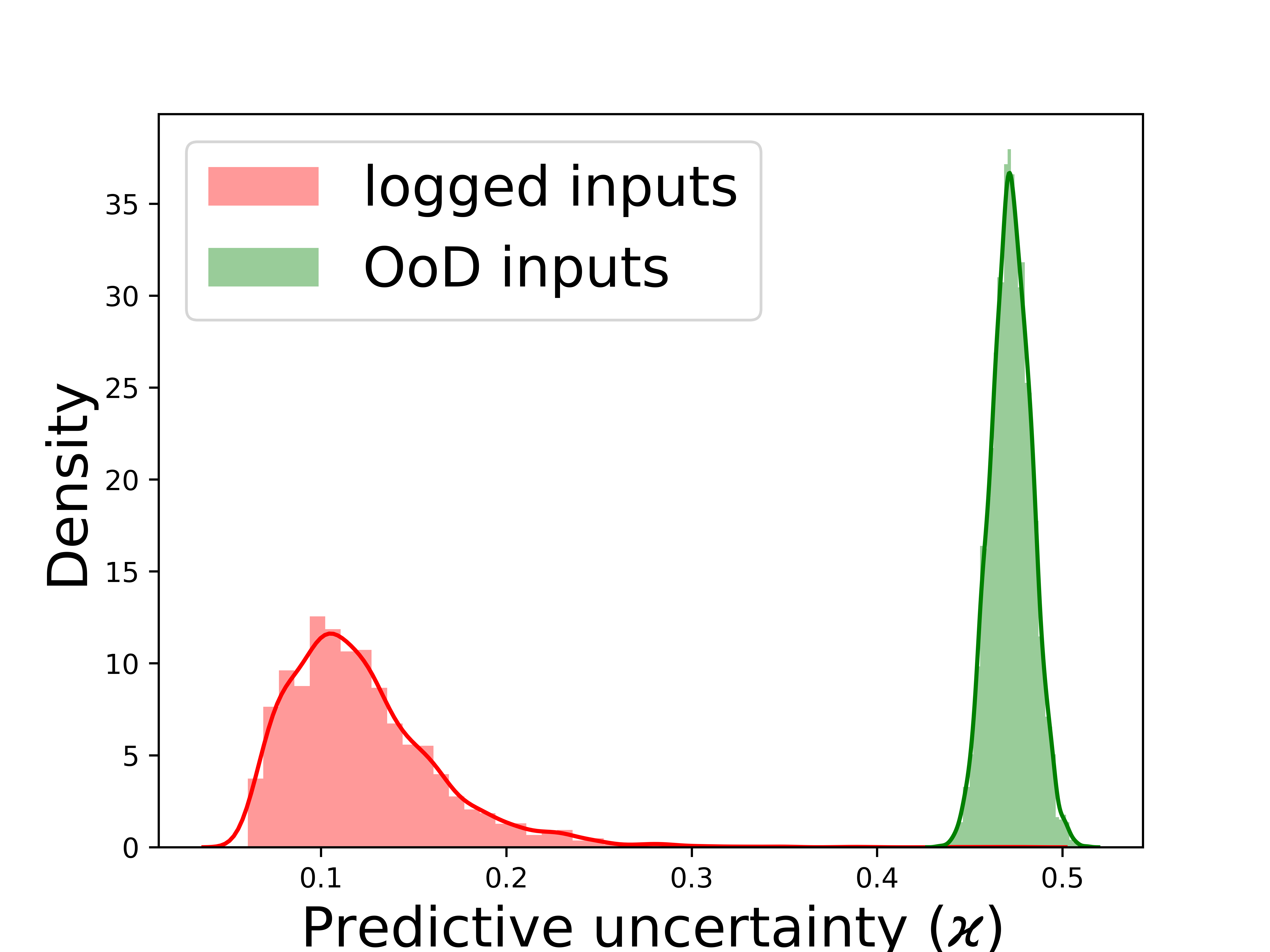}\label{fig:hist_norm}}
\subfloat[]{\includegraphics[width=.15\textwidth]{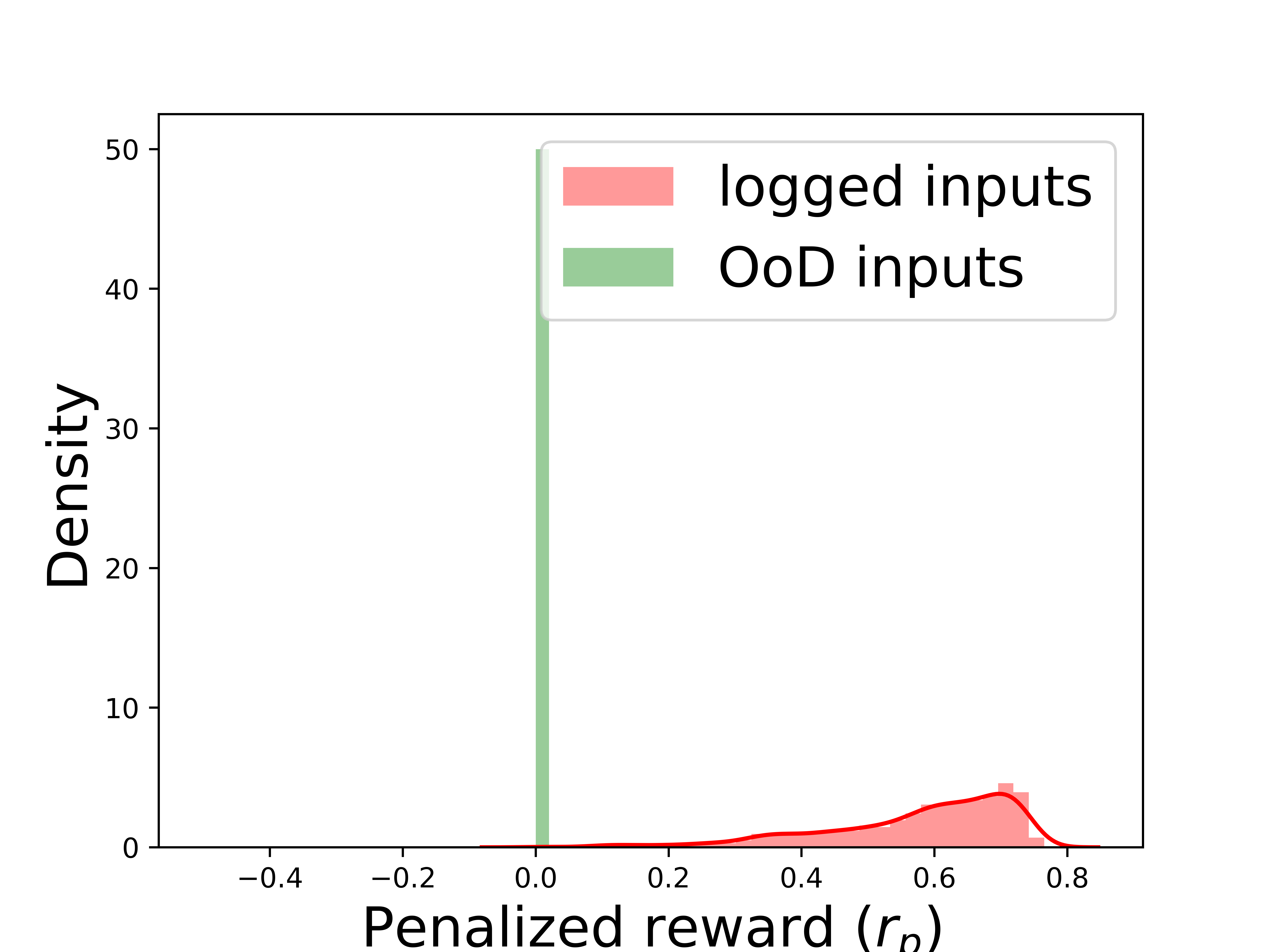}\label{fig:hist_rewards}}
\caption{(a,b,c): the distribution of $\kappa$, $\varkappa$ and $r_p$ for the logged inputs compared with the counterpart for the OoD inputs with only control parameter values are randomly generated; (d,e,f): the distribution of $\kappa$, $\varkappa$ and $r_p$ for the logged inputs compared with the counterpart for the OoD inputs whose conditional and control parameters are all randomly generated.}
\label{fig:qualitative2}
\end{figure}

To answer the second question, we evaluate the distribution discrepancy ($\kappa$), the predictive uncertainty ($\varkappa$) and the penalized rewards ($r_p$) based on the generated control results by the ensemble of cGANs given three set of inputs: 1) the first set is the logged conditional and control parameters in the testing data; 2) the second set is the logged conditional parameters in the testing data, but combined with randomly generated control parameters to mimic OoD inputs; 3) the third set consists of OoD inputs whose conditional and control parameters are all randomly generated. We compare the empirical distribution of $\kappa$, $\varkappa$ and $r_p$ for the logged inputs and the OoD inputs in the second set in Figure~\ref{fig:hist_divs2}, \ref{fig:hist_norms2} and \ref{fig:hist_rewards2} respectively. Furthermore, we compare the empirical distribution of $\kappa$, $\varkappa$ and $r_p$ for the logged inputs and the OoD inputs in the third set in Figure~\ref{fig:hist_kappa}, \ref{fig:hist_norm} and \ref{fig:hist_rewards} respectively. As can be seen, when given OoD inputs with only control parameters are randomly generated, we cannot perfectly distinguish OoD inputs from logged inputs using predictive uncertainty of control results (Figure~\ref{fig:hist_norms2}), but can distinguish easily using discrepancy between predicted distributions of control results (Figure~\ref{fig:hist_divs2}). When given OoD inputs whose conditional and control parameters are all randomly generated, we cannot perfectly distinguish OoD inputs from logged inputs using discrepancy between predicted distributions of control results (Figure~\ref{fig:hist_kappa}), but can distinguish easily using predictive uncertainty (Figure~\ref{fig:hist_norm}). However, we can easily distinguish OoD inputs from logged inputs using our epistemic-uncertainty-penalized rewards in both scenarios (Figure~\ref{fig:hist_rewards2} and \ref{fig:hist_rewards}). From the above results, we can conclude that to accurately avoid giving over-estimated rewards to OoD inputs, it is necessary to use both the discrepancy between predicted distributions and the amount of predictive uncertainty to penalize rewards.

\section{Experiment on Continuous Control}
In this section, we present the experiment in which we applied our proposed framework on the industrial benchmark (IB), a public available simulator with properties inspired by real industrial systems with continuous control \cite{hein2017benchmark}. The process of searching for an optimal control policy on the IB is to resemble the task of finding optimal valve settings for gas turbines or optimal pitch angles and rotor speeds for wind turbines.

\begin{algorithm}[t]
  \caption{Offline model-based optimization instantiation with DDPG for continuous control policy learning}
  \begin{algorithmic}[1]
  	\REQUIRE A historical dataset $\mathcal{D}$, the reward function $r(\mathbf{y})$
  	\STATE  Train on the historical dataset $\mathcal{D}$ an ensemble of $M$ cGAN models, $\{G_{i} \}_{i=1}^M$ 
  	\STATE Randomly initialize the critic network $Q(\mathbf{x},\mathbf{u})$ and policy (actor) network $\pi(\mathbf{x})$  
  	\STATE Initialize target network $Q'(\mathbf{x},\mathbf{u})$ and $\pi'(\mathbf{x})$ with same weights with $Q(\mathbf{x},\mathbf{u})$ and $\pi(\mathbf{x})$ respectively
  	\STATE Initialize replay buffer $R$
  	\FOR{episode=1,K} 
  	\STATE {Initialize a random process $\mathcal{N}$ for action exploration} 
  	\STATE {Sample an initial conditional parameters $\mathbf{x}_0$ from $\mathcal{D}$} 
  	\FOR{$t=1,T$}
  	\STATE {Select control parameters $\mathbf{u}_t=\pi(\mathbf{x}_t)+\mathcal{N}_t$ according to the current policy and exploration noise}
  	\FOR{$i=1,M$}
  	    \STATE {Generate $N$ control result samples by $\tilde{\mathbf{y}}_i^j = G_{i}(\mathbf{z}_j|\mathbf{x}_t,\mathbf{u}_t)$ where $\mathbf{z}_j \sim p(\mathbf{z})$, $j=1,2,...,N$}
  	\ENDFOR
  	\STATE { Compute the epistemic-uncertainty-penalized reward $r_p(\mathbf{x}_t,\mathbf{u}_t)$.}
  	\STATE {Randomly pick a sample $\mathbf{y}_t$ from $\{ \tilde{\mathbf{y}}_i^j \}_{i=1:M}^{j=1:N}$ and set $\mathbf{x}_{t+1}=\mathbf{y}_t$}.
  	\STATE {Add sample $\big(\mathbf{x}_t,\mathbf{u}_t,r_p(\mathbf{x}_t,\mathbf{u}_t),\mathbf{x}_{t+1} \big)$ to $R$}
  	\STATE {Use DDPG to update $Q,\pi$ and $Q',\pi'$}
  	\ENDFOR
  	\ENDFOR
  \end{algorithmic}
   \label{alg:rl}
\end{algorithm}

Control policies in the IB need to specify changes $\Delta v$, $\Delta g$ and $\Delta s$ in three observable steering variables $v$ (velocity), $g$ (gain) and $s$ (shift). In our experiment, we set the conditional parameters at a time $t$ as $\mathbf{x}_t=[v_t,g_t,s_t, c_{t}, f_{t}]$ where $c_{t} \geq 0$ (consumption) and $f_{t} \geq 0$ (fatigue) are two reward relevant variables; the control parameters at time $t$ as $\mathbf{u}_t=[\Delta v_t,\Delta g_t,\Delta s_t]$; the control results of time $t$ as $\mathbf{y}_t=[v_{t+1}, g_{t+1}, s_{t+1}, c_{t+1}, f_{t+1}]$. The reward of time $t$ is calculated as $r(\mathbf{y}_t)=-c_{t+1}-3f_{t+1}$.

We generate two sets of historical data using two behaviour policies: a random behaviour policy and a safe behaviour policy. Specifically, the random behaviour policy will set control parameters which are randomly sampled from the value space. The safe behaviour policy will keep the velocity and gain only in a medium range using a conservative strategy such that a very limited value space for velocity and gain will be covered. We use each behaviour policy to simulate 300 trajectories of length 100 as the historical data. We train a cGAN ensemble-based surrogate model from each historical dataset, and then use DDPG \cite{lillicrap2016continuous} as the backbone of the underlying RL algorithm to learn a continuous control policy. The implementation details are given in Algorithm~\ref{alg:rl}. Using the same DDPG backbone, we also compare the performance of our learned policies with same baselines in the previous section. To get a fair comparison, on each historical dataset we run the training algorithm for 1500 episodes with 5 random seeds for each method. We report the undiscounted reward of the selected policies on the IB simulator averaged over the last 100 training episodes with the 5 random seeds, $\pm$ standard deviation. The results are reported in Table~\ref{tab:mopo} in which the performance of the behaviour policies is also included. As can be seen, our proposed method achieves the highest average reward on both scenarios with a clear margin. We also find that our proposed method is the only one which can achieve better performance than the behaviour policy in both scenarios.

\begin{table}
\caption{The performance of the behaviour policy and control policies learned from historical data by different methods on IB.}
\label{tab:mopo}
\begin{center}
\begin{tabular}{lcc}
 &  Random & Safe  \\
 \hline
behaviour policy & -247.0 $\pm$ 32.8 & -216.3 $\pm$ 12.0 \\
Baseline1 &  -254.2 $\pm$ 47.4 &  -436.3 $\pm$ 54.9\\
Baseline2 &  -270.1 $\pm$ 25.9&  -319.8 $\pm$ 34.3\\
Baseline3 & -258.4 $\pm$ 21.7 & -344.4 $\pm$ 10.3 \\
Baseline4 & -239.0 $\pm$ 15.4 & -222.8 $\pm$ 10.4 \\
Our method & \textbf{-207.3} $\pm$ 13.1& \textbf{-201.1} $\pm$ 11.2 \\
 \hline
\end{tabular}
\end{center}
\end{table}

\section{Related Work}
Learning optimal control policies from offline data is known to suffer from the distribution shift problem. Several techniques have been proposed to tackle the problem. For example, one approach is to constrain the learned policy to be closer to the behaviour policy which is adopted by many model-free offline RL algorithms \cite{levine2020offline}.
Another approach is to train a conservative dynamics model or critic that minimizes the predictions on OoD inputs for offline model-based optimization \cite{trabucco2021conservative,yu2021combo}. A more common approach for offline model-based optimization is to utilize uncertainty quantification to evaluate the risk of giving overestimated rewards to OoD inputs \cite{yu2020mopo,kidambi2020morel}. Our work also takes the uncertainty-quantification approach, but with the difference that we leverage ensemble of cGANs as the probabilistic model to produce more realistic control result samples for reliable uncertainty quantification.

Data-driven control policy learning have been applied in industrial systems for more than decades \cite{schlang1999neural}. However, most of them focus on online scenarios where there is a simulator of the industrial process to provide control feedback \cite{nian2020review}. In practice, building accurate simulation models for industrial systems is often very costly or infeasible. Learning control policies for industrial processes from offline data has been recently explored. For example, a particle swarm optimization-based algorithm is combined with a planning-based strategy for offline industrial control policy learning~\cite{hein2017batch}. A generic programming-based approach is proposed to generate interpretable control policies for industrial systems from offline data \cite{hein2018interpretable}. Similar to our work is \cite{depeweg2018decomposition} in which a risk-sensitive model-based offline RL algorithm based on Bayesian neural networks is proposed to handle noisy industrial systems. Different from previous approaches, our proposed surrogate model is general to both the discrete and continuous control cases, and is agnostic to the optimization algorithm being used.

\section{Conclusion}
In this work, we extend the current success of offline model-based optimization to real-world industrial control problems by introducing a cGAN ensemble-based uncertainty-aware surrogate model. Through the experiments on two industrial control problems, we show the proposed surrogate model can reliably improve performance of offline learned control policies for noisy industrial processes.

\bibliographystyle{IEEEtran}
\bibliography{icmlbib}

\end{document}